# Application Layer Definition and Analyses of Controller Area Network Bus for Wire Harness Assembly Machine

Hui Guo, Ying Jiang
*Institute of Mechanics and Automatization, Shanghai University*
*Email: gauveymail@gmail.com*

## Abstract

*With the feature of multi-master bus access, nondestructive contention-based arbitration and flexible configuration, Controller Area Network (CAN) bus is applied into the control system of Wire Harness Assembly Machine (WHAM). To accomplish desired goal, the specific features of the CAN bus is analyzed by compared with other field buses and the functional performances in the CAN bus system of WHAM is discussed. Then the application layer planning of CAN bus for dynamic priority is presented. The critical issue for the use of CAN bus system in WHAM is the data transfer rate between different nodes. So processing efficient model is introduced to assist analyzing data transfer procedure. Through the model, it is convenient to verify the real time feature of the CAN bus system in WHAM.*

## 1. Introduction

As the assembly of wire harness is one of the most difficult and time consuming assembly phase, Wire Harness Assembly Machine (WHAM) plays more and more prominent role in the field of electronics [1]. WHAM is built under stringent machine structure constrains and tightly attached to quality-checking equipments. Difficulty of system Extending and transplant are found by means of PCI (Peripheral Component Interface) bus motion controller or PC (Programmable Controller). While field bus technology with reconfiguration and extending develops rapidly, so field bus is under our consideration for its wildly successful application in many domains. Complexity reduction for the control system requires appropriate bus technology which can provide an efficient and real-time communication between the different devices with maximum freedom for reconfiguration.

The applicability of CAN bus technology for real-time communication is investigated, as the WHAM will be working at high speed. The underlying data handling in the CAN bus system has to deal with mass data originating from several independent parts. To prove the feasibility of CAN bus in WHAM, processing efficiency model is presented by analyzing CAN specification.

Field buses are digital networks that provide an efficient digital serial data exchange among one or more controllers and a number of field devices. Real-time is the main requirement in the control system of WHAM. But there are no standard available, and the user needs tools to guide the process of choosing the field bus best suited for WHAM. Comparative analyses are presented concerning the performance of four field buses which is widely used in electronics machines: MODBUS PLUS, PROFIBUS-PA, and CAN (see Table 1). Throughput and responsiveness in PROFIBUS-PA and MODBUS PLUS depends on token rotation time. The standard of MODBUS PLUS is proprietary, and using it would be bound to corresponding limits and difficult to be further expanded. PROFIBUS-PA permitting only one master node is difficult for the realization the multi-master control system. CAN bus is only the one that becomes the standard of International Standard Organization, which means that CAN will be easily to get detailed content of the bus working mechanism and the off-the-shelf devices including built-in interfaces for CAN bus. Due to its source from automotive applications, the CAN bus is very promising candidate in terms of functionality, performance, environmental robustness and cost.

## 2. Applicability analyses for CAN bus technology in WHAM

To apply the CAN bus into the controlling of wire harness assembly, it is necessary to analyze the specification of CAN. Firstly the availability of CAN bus is discussed, and then the mechanism of data

| Name | Standard | Master | Max. Speed | Topology | Max. Length (m @Max. bit rate) | Max. data / Protocol Data Unit | Data Link Layer Implementation |
|---|---|---|---|---|---|---|---|
| MODBUS PLUS | proprietary | multi | 1Mbps | Bus | 1800 | 32bytes | chip |
| PROFIBUS-PA | EN50170 | single | 93.75Kbps | bus | 1900 | 246bytes | ASIC |
| CAN | ISO11898, ISO11519 | multi | 1Mbps | bus | 40 | 8bytes | chip |

**Table 1. Comparisons of three field buses [2], [3], [4].**

transfer and the elaborate things is introduced regarding data coding and bus accessing.

## 2.1. Availability and standardization

CAN is a serial communication protocol which efficiently supports distributed real-time control with a very high level of security [5]. Furthermore it efficiently supports functional redundancy due to its multi-master capabilities. The success of CAN bus in automotive industry, together with its low cost and high performance has led to rapidly increasing application in many other control field. For example, some recent commercial combine harvesters are equipped with a CAN bus control system [6]. With regard to customers' demands and mechanical robustness, CAN bus is suitable for the task-allocating in the control system of WHAM.

| LAYER | CAN Equivalent LAYER |
|---|---|
| LAYER 7 APPLICATION | User's definition |
| LAYER 6 PRESENTATION | Not explicitly defined |
| LAYER 5 SESSION | Time-Triggered Communication |
| LAYER 4 TRANSPORT | Not explicitly defined |
| LAYER 3 NETWORK | Not explicitly defined |
| LAYER 2 DATA LINK | Logic Link Control |
| | Medium Access Control |
| LAYER 1 PHYSICAL | Explicitly defined |
| LAYER 0 TRANSMISSION MEDIUM | |

**Figure 1. OSI/RM model.**

According to OSI/RM (Open System Interconnect/ Reference Model) the CAN architecture is constituted by two layers: DLL (Data Link Layer) and PL (Physical Layer) (see Fig. 1). The first and second layers are international standards, so the defined interface to second layer for our application is employed. On this basis some software development environment have available to test functionality of CAN bus, which makes the fast development and cost reduction possible.

## 2.2. Data transferring and handling

Although the total length of CAN bus system must be no longer than forty meters operating at the maximum data rate of 1 Mbps, the length is sufficient for WHAM in which all executive nodes sensor nodes and indicator nodes are arranged in a small space (the space dimension of WHAM is 1.2 meters long, 0.8 meters wide and 1.5 meters high). The CAN bus network consists of a twisted pair wires, and different devices can be connected to the CAN bus network and communicate with each other by exchanging messages. Physically, a message consists of a sequence of bits. According to expected processing speed of WHAM, we can acquire that the transfer time of messages across the network should be less than 1 second to process one wire harness. In the next paragraph, we introduce processing efficiency model which is help to prove the feasibility of CAN bus used in WHAM.

## 2.3. Arbitration mechanism

A rigid Master-Slave medium access mechanism is not feasible for the control system of WHAM, for necessary redundancy of the bus master is required according to the working mechanism of WHAM. CAN bus uses a multi-master, priority-based bus access with a modified CSMA/CD access method and all nodes are operating as masters, which means that all nodes are able to transmit data and multiple nodes can request access to the bus simultaneously. Any node may start to transmit a frame when the bus is idle. If two or more nodes start to transmit frames at the same time, the bus access contention is resolved by bit-wise arbitration using arbitration field which is 11 bits identifier in standard format [5]. The method of arbitration guarantees that no information and no time are lost. The frame with the highest priority dominates the bus until this period of conversion is over between a transmitter and a receiver. So the protocol realizes inherently flexible communication structures and broadcasting capabilities. Frames that have lost arbitration will participate in a new arbitration round when the bus is idle again. The resulting time delay depends on the other higher priority frames. So time delay of the highest priority frame is predictable. Nodes may be added to the CAN network without requiring any changes in the software or hardware. It is very convenient to connect expanding electronic parts to WHAM.

| Frame type | DATA | REMOTE | ERROR | INTER-FRAME |
|---|---|---|---|---|
| Start of Frame | 1 | 1 | No | No |
| Arbitration Frame | 12 | 12 | No | No |
| Control Field | 6 | 6 | No | No |
| Data Field | 0~8 | 0 | No | No |
| CRC Field | 16 | 16 | No | No |
| ACK Field | 2 | 2 | No | No |
| End of Frame | 7 | 7 | No | No |
| Error Flag | No | No | 6~12 | No |
| Error Delimiter | No | No | 8 | No |
| Intermission | No | No | No | 3 |
| Suspend Intermission | No | No | No | 8 |
| Bus idle | No | No | No | n |
| Total | 44~52 | 44 | 14~18 | 11+n |

Note: n∈N

**Table 2. Frame length of standard format.**

### 2.4. Length of four different types frames

In the CAN bus system, message transfer is maintained and controlled by four different frame types: DATA frame, REMOTE frame, ERROR frame, and OVERLOAD frame. After analyzing the frame structure in CAN specification 2.0, we can get the message length in each frame (see Table 2).

## 3. Application layer planning of CAN bus system

From the former paragraph we know that CAN bus is based on priority, so logically configuration according working flow of WHAM is critical for development of the CAN bus system.

### 3.1. Working mechanism of WHAM

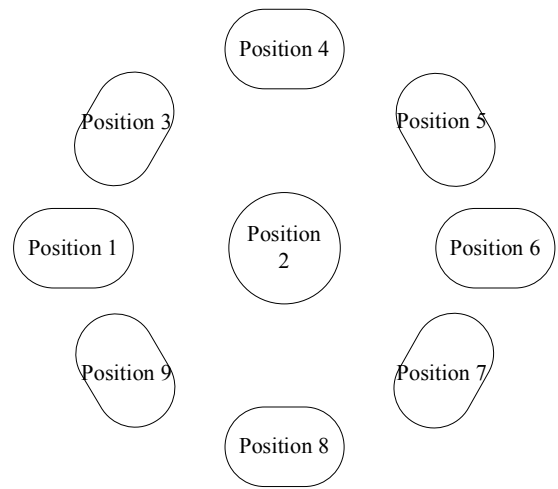

**Figure 2. Layout of working position in WHAM.**

According to the working flow, we can draw the working-position graph showing in Fig. 2. In each position there are executors, detectors or indicators which are connected to CAN bus system as a node. Each working position has its own unique function name (see Table 3), and they fulfill the expected task by cooperating with each other.

### 3.2. Node identifier planning in CAN bus system for dynamic priority

The value of node identifier in a CAN message determines the priority of the message. To achieve dynamic priority for all CAN bus nodes, 11-bits can be segmented in the following format: ###****XXXX.

The first three bits represent operation modes of the CAN bus system. WHAM may work at four states: normal mode, error mode, customized mode and self-check mode. At any given time WHAM is working at only one mode, so the first three bits of all identifiers should be same. Through such coding method we can easily trace the error nodes.

| Frame Identifier (Binary) | Description |
|---|---|
| 001****XXXX | Normal mode (mode 1) |
| 010****XXXX | Customized mode (mode 2) |
| 011****XXXX | Error mode (mode 3) |
| 100****XXXX | Self-check mode (mode 4) |
| ###0000XXXX | WHAM halts |
| ###0001XXXX | Error found by detector 1 |
| ###0010XXXX | Error found by detector 2 |
| ###0011XXXX | Error found by detector 3 |
| ###0100XXXX | Error found by detector 4 |
| ###1111XXXX | No error |
| ###****0001 | Display node |
| ###****0010 | Executor 1 |
| ###****0011 | Executor 2 |
| ###****0100 | Executor 3 |
| ###****0101 | Detector 1 |
| ###****0110 | Executor 4 |
| ###****0111 | Executor 5 |
| ###****1000 | Detector 2 |
| ###****1001 | Executor 6 |
| ###****1010 | Detector 3 |
| ###****1011 | Executor 7 |
| ###****1100 | Detector 4 |
| ###****1101 | Executor 8 |
| ###****1110 | Executor 9 |
| ###****1111 | Executor 10 |

**Table 3. Bits assignment of CAN bus frame identifier.**

The middle four bits are used for error mark identification. CAN bus system can have one of two complementary logical values: 'dominant' representing 0 and 'recessive' representing 1. During simultaneous transmission of 'dominant' and 'recessive' bits, the result of CAN bus value will be 'dominant'. The middle four bits are all recessive, demonstrating that WHAM is working at the setting mode. Whenever errors are found, a message with dominant bit in the middle part of the identifier can dominate the CAN bus.

More occurrences of ERROR frame may reduce the processing efficiency or cause lots of inferior wire harness. In the application layer, $w_i$ is defined as occurrences times of the CAN bus node ERROR frame. When the value of $w_i$ is more than ten, WHAM will be halted by the highest priority frame from Display node.

The last four bits are used to distinguish different CAN bus nodes, and identifiers are planned according to working flow (see Table 3).

### 3.3. Processing efficiency model of WHAM

$$\textbf{min.}\ T(x,y) = \sum_{i=1}^{3} T_i(x,y) \quad (1)$$

$$T_1(x,y) = \sum_{i=1}^{n}(1+m_i)[1+P_i(x,y)]B_i(y)M_i$$

$$T_2(x,y) = \sum_{i=1}^{n} w_i B_i(y)E_i + \sum_{i=1}^{n} B_i(y)I_i$$

$$T_3(x,y) = C(x,y)$$

**s.t.**

$$P_i(x,y) = \max \sum_{j=1}^{r} x(j)/r\big|_y$$

$$B_i(y) = (1+p+ps1+ps2+pr) \times pe \times q$$

$$M_i \in [14,18] \cup [44,52]$$

$$E_i \in [0,10]$$

$$I_i \in \{3n+m \mid n \in N \cap n \le 3, m \in N \cap m \le 10\}$$

$$C(x,y) = Syn(x,y)L_i$$

$$x \le 10$$

$$y = 1,\ 2,\ 3,\ 4$$

$$m_i \in [0,10]$$

$$w_i \in [0,10]$$

$$r = 100$$

$$ps1 = 1,\ 2,\ 3,\ 4,\ 5,\ 6,\ 7,\ 8$$

$$ps2 = 1,\ 2,\ 3,\ 4,\ 5,\ 6,\ 7,\ 8$$

$$pr = 0,\ 1,\ 2$$

The efficiency of the WHAM is a critical factor when designing the control system and it will be helpful to know the minimum processing time; thereby processing efficiency model is introduced by applying optimization theory and statistic method (see (1)). The parameters and functions are defined as follows:

$T(x,y)$    is the one circle time function of data transfer in WHAM.

$x(j)$    denotes the times of a CAN bus node transferring error data frame at the $j$

| | |
|---|---|
| $y$ | processing circle. $x(j) \in N$. denotes the code of WHAM working mode. $y(j) \in N \cap y \leq 4$ (see Table 4). |
| $P_i(x, y)$ | is the power function of the number $i$ CAN bus node at the $y$ mode. To avoiding disqualified product appearing, adaptive method is introduced. In our program a counter is used in every CAN bus node to store the error times of DATA frame, which is initialed by different number according to WHAM mechanism (see Table 6). |
| $j$ | denotes the circle times of data transfer in CAN bus system. $j \in N$. |
| $r$ | denotes the maximum value of $j$, which is preferred to initial 100, for a bundle of wire harness is produced through 100 data transfer circles. |
| $B_i(y)$ | is one nominal bit time function of the number $i$ CAN bus node at the $y$ mode. |
| $p$, $ps1$, $ps2$ and $pr$ | are all parameters of NOMINAL BIT TIME [5]. |
| $p$ | denotes the time quanta of PROP_SEG [5]. |
| $ps1$ | denotes the time quanta of PHASE_SEG1 [5]. |
| $ps2$ | denotes the time quanta of PHASE_SEG2 [5]. |
| $pr$ | denotes the time quanta of INFORMATION PROCESSING TIME which is programmable [5]. |
| $q$ | denotes the time quantum which is a fixed unit of time derived from the oscillator period [5]. |
| $pe$ | denotes the value of the prescaler which is programmable. As $pe$ has close relation with hardware rather than software, we regard it as a fixed value to simplify our model. |
| $M_i$ | denotes the DATA frame length transmitted by the number $i$ CAN bus node. |
| $m_i$ | denotes the times of DATA frame retransmitted by the number $i$ CAN bus node. |
| $E_i$ | denotes the ERROR frame length transmitted by the number $i$ CAN bus node. |
| $w_i$ | denotes the times of ERROR frame transferred by the number $i$ CAN bus node. |
| $I(i, i+1)$ | is the function that represents the length of INTERFRAME. |
| $C(x, y)$ | is a correction function that corrects function $T(x, y)$, for data transfer may be accelerated or delayed by various tiny changes of the CAN bus system. Compared with $T_1$ and $T_2$ and, $C(x, y)$ has little weight in $T(x, y)$, herewith statistic method is introduced to give the approximate value. |
| $Syn(x, y)$ | denotes the CAN bus synchronization time. |
| $L_i$ | denotes the distance between the i CAN bus node behaving as the transmitter, and the node behaving as receiver. |

| $y$ | Mode |
|---|---|
| 1 | Normal mode (mode 1) |
| 2 | Customized mode (mode 2) |
| 3 | Error mode (mode 3) |
| 4 | Self-check mode (mode 4) |

**Table 4. Definition of *y*.**

## 4. Conclusions

Through above analyses, conclusion can be drawn that the CAN bus is very promising candidate for the WHAM application due to its basic concepts: multi-master priority bus access, nondestructive contention-based arbitration, and configuration flexibility. As the discussion above, the application layer is defined in CAN bus by users, thereupon CAN bus application layer planning for dynamic priority is presented. With the planning, it is convenient to find out where the error node appeared and improve the efficiency of data-filtering. The vital issue for the use of CAN bus system in WHAM is whether CAN bus has the ability of transferring mass data in brief time between different nodes, whereupon processing efficiency model is introduced to give the theory support of the feasibility of CAN bus in WHAM. Furthermore, with the guide of processing efficiency model, it becomes convenient to find out the possible data transfer bottle-neck and time waste phases in the CAN bus system of WHAM.

## References

[1] E.Aguirre and B.Raucent, "Performance of Wire Harness Assembly Systems", proceedings of ISIE'94, Santiago, May 1994, pp. 292-297.

[2] http://www.fieldbus.org/.

[3] J.Klaus and B.Annerose, "Application of Industrial CAN Bus Technology for LEO-satellites," Acta Astronautica, vol. 46, pp. 313-317.

[4] S. J. Luis, P. Arturo and J. Gabriel, "Analysis of Channel Utilization for Controller Area Networks," Computer Communications, vol. 21, pp. 1446-1451.

[5] Robert Bosch GmbH, CAN Specification, Version 2.0, Sep. 1991.

[6] G.Craessaerts, K.Maertens, J. and De Baerdemaeker, "A Windows-based Design Environment for Combine Automation via CANbus," Computers and Electronics in Agriculture, vol. 49, pp. 233-245.